\documentclass[11pt,a4paper]{article}
\usepackage{emnlp2021}
\usepackage{times}
\usepackage{latexsym}

\usepackage{microtype}

\usepackage{mymathdef}

\usepackage{graphicx}
\usepackage{amsmath}
\usepackage{amsthm}
\usepackage{booktabs}
\usepackage{algorithm}
\usepackage[noend]{algorithmic}
\usepackage{booktabs}
\usepackage{multirow}
\usepackage{threeparttable}
\usepackage{float}
\usepackage{xspace}

\usepackage{tikz}
\usepackage{pgfplots}
\usepackage{caption}
\usepackage{subcaption}

\title{Keyphrase Generation with Fine-Grained Evaluation-Guided Reinforcement Learning}

\author{Yichao Luo\thanks{\ \  These two authors contributed equally.}, Yige Xu\footnotemark[1], Jiacheng Ye, Xipeng Qiu, Qi Zhang\thanks{\ \ Corresponding author.}\\
  Shanghai Key Laboratory of Intelligent Information Processing, Fudan University \\
  School of Computer Science, Fudan University \\
  Songhu Road 2005, Shanghai, China
 \\
  {\tt \{ycluo18,ygxu18,yejc19,xpqiu,qz\}@fudan.edu.cn} \\}

\date{}

\begin{document}
\maketitle
\begin{abstract}
Aiming to generate a set of keyphrases, Keyphrase Generation (KG) is a classical task for capturing the central idea from a given document.
Based on Seq2Seq models, the previous reinforcement learning framework on KG tasks utilizes the evaluation metrics to further improve the well-trained neural models.
However, these KG evaluation metrics such as $F_1@5$ and $F_1@M$ are only aware of the exact correctness of predictions on phrase-level and ignore the semantic similarities between similar predictions and targets, which inhibits the model from learning deep linguistic patterns.
In response to this problem, we propose a new fine-grained evaluation metric to improve the RL framework, which considers different granularities: token-level $F_1$ score, edit distance, duplication, and prediction quantities.
On the whole, the new framework includes two reward functions: the fine-grained evaluation score and the vanilla $F_1$ score.
This framework helps the model identifying some partial match phrases which can be further optimized as the exact match ones.
Experiments on KG benchmarks show that our proposed training framework outperforms the previous RL training frameworks among all evaluation scores. In addition, our method can effectively ease the synonym problem and generate a higher quality prediction.
The source code is available at \url{https://github.com/xuyige/FGRL4KG}.

\end{abstract}

\section{Introduction}
\label{sec-intro}

Keyphrase Generation (KG) is a classical but challenging task in Natural Language Processing (NLP), which requires automatically generating a set of keyphrases. Keyphrases are short phrases that summarized the given document. Because of the condensed expression, keyphrases can be beneficial to various downstream tasks such as information retrieval~\cite{jones1999phrasier}, opinion mining~\cite{wilson-etal-2005-recognizing,berend-2011-opinion}, document clustering~\cite{DBLP:conf/acl/HulthM06}, and text summarization~\cite{wang-cardie-2013-domain}.

In recent years, end to end neural models have been widely-used in generating both present and absent keyphrases. ~\citet{DBLP:conf/acl/MengZHHBC17} introduced CopyRNN, which consists of an attentional encoder-decoder model~\cite{DBLP:conf/emnlp/LuongPM15} and a copy mechanism~\cite{DBLP:conf/acl/GuLLL16}. After that, relevant works are mainly based on the sequence-to-sequence framework~\cite{DBLP:conf/acl/YuanWMTBHT20,DBLP:conf/emnlp/ChenZ0YL18,DBLP:conf/aaai/ChenGZKL19}. Meanwhile, $F_1@5$~\cite{DBLP:conf/acl/MengZHHBC17} and $F_1@M$~\cite{DBLP:conf/acl/YuanWMTBHT20} are used for evaluating the model prediction. $F_1@5$ computes the $F_1$ score with the first five predicted phrases (if the number of phrases is less than five, it will randomly append until there are five phrases). $F_1@M$ compares all keyphrases (variable number) predicted by the model with the ground truth to compute an $F_1$ score. Furthermore, \citet{DBLP:conf/acl/ChanCWK19} utilize the evaluation scores as the reward function to further optimize the neural model throughout the reinforcement learning (RL) approach.

However, the traditional $F_1$-like metrics are on phrase-level, which can hardly recognize some partial match predictions. For example, supposing that there is a keyphrase called ``{\it natural language processing}'', and one model provides a prediction called ``{\it natural language generation}'' while another model provides ``{\it apple tree}''. Both of these two phrases will get zero score from either $F_1@5$ or $F_1@M$. But it is undoubtedly that ``{\it natural language generation}'' should be a better prediction than ``{\it apple tree}''. \citet{DBLP:conf/acl/ChanCWK19} propose a method to evaluate similar words, but they only consider abbreviations of keywords and use it only during the evaluation stage.

In response to this problem, we propose a \textbf{F}ine-\textbf{G}rained ($FG$) evaluation score to distinguish these partial match predictions. First, in order to align the $F_1$ score, the exact correct predictions will obtain the $FG$ score of one (e.g., {\it natural language processing} mentioned above), and the absolutely incorrect predictions will obtain the $FG$ score of zero (e.g., {\it apple tree} mentioned above). Second, for partial match predictions like ``{\it natural language generation}'', $FG$ score, our proposed metric, will compare the prediction with the target in the following perspectives:(1) prediction orders in token-level; (2) prediction qualities in token-level; (3) prediction diversity in instance-level; (4) prediction numbers in instance-level. The specific detail of our proposed $FG$ score can be seen in Section~\ref{sec-fg-score}.

\begin{figure}
  \centering

  \tikzstyle{edge}=[-latex',thick,draw=black!90,shorten <=1pt,shorten >=1pt]
\tikzstyle{block}=[draw, text width=5em,align=center,shape=rectangle, rounded corners, , align=center]
  \tikzstyle{nobox}=[align=center]

\begin{tikzpicture}[node distance=6em,auto,font=\footnotesize\selectfont]

    \node [block,text width=3em,] (bert) {Encoder-Decoder Model};
    \node [block, text width=4em,] (x1) [right of=bert] {RL w/ $FG$ Score};

    \node [block] (x2) [right of=x1,node distance=3.25em,yshift=4em]{RL w/ $F_1$ Score};
    \node [block] (x3) [right of=x1,node distance=3.25em,,yshift=-4em] {RL w/ $FB$ Score};
    \node [block] (x4) [right of=x3,node distance=6em,,yshift=4em] {Final Model};

    \path[edge,] (bert) |-  (x2);
    \path[edge,] (x2) -|  (x4);
    \path[edge,densely dashed,,blue,] (x2) -- (x4.172);
    \path[edge,red,densely dotted] (x2) -- (x4.135);

    \path[edge,densely dashed,,blue,] (bert.10) --  (x1.172);
    \path[edge,densely dashed,,blue,] (x1) -- (x2);

    \path[edge,densely dotted,green!70!black] (x1) -- (x3);
    \path[edge,red,densely dotted] (x3) -- (x2);
    \path[edge,red,densely dotted] (bert) |-  (x3);
\end{tikzpicture}
	  \caption{Flow chart of three reinforcement learning methods. The blue edges and red edges are our proposed reinforcement learning methods (catSeq*+2RL($FG$) and catSeq*+2RL($FB$)). The green densely dotted line means the $FB$ score. We use some data generated by $FG$ score to train the BERT model, and the BERT model is used to compute the $FB$ score.}\label{fig:flow-chart}
\end{figure}

Based on previous works that use the reinforcement learning technique and adopt the self-critical policy gradient method~\cite{DBLP:conf/cvpr/RennieMMRG17}, we propose a two-stage RL training framework for better utilizing the advantages of $FG$ score. As shown in Figure~\ref{fig:flow-chart}, the black edges show the previous RL process and the blue edges show our proposed two-stage RL process. Our two-stage RL can be divided into two parts: (1) First, we set $FG$ score as the adaptive RL reward and use RL technique to train the model; (2) Second, we use $F_1$ score as the reward, which is the same as \citet{DBLP:conf/acl/ChanCWK19}. Furthermore, in order to make $FG$ score smoothly, we carefully train a BERT~\cite{devlin-etal-2019-bert} model to expand the original $FG$ score from discrete to continuous numbers (the green line in Figure~\ref{fig:flow-chart}). This BERT scorer can predict a continuous $FG$ score, which can also be used in our two-stage RL framework (the red edges in Figure~\ref{fig:flow-chart}).

Comparing with the $F_1$ score, our $FG$ score has two main advantages: (1) $FG$ score can recognize some partial match predictions, which can better evaluate the quality of predictions in a fine-grained dimension; (2) During the reinforcement learning stage, $FG$ score can provide a positive reward to the model if it predicts some partial match predictions, while the $F_1$ score will return a negative reward of zero in this situation. Therefore, in our proposed two-stage RL framework, the first stage can help the model predict some partial match phrases, and the second stage can further promote the partial match phrases to the exact match phrases. We conduct exhaustive experiments on keyphrase generation benchmarks and the results show that our proposed method can help better generating keyphrases by improving both the traditional $F_1$ score and the $FG$ score. In addition to this, we also conduct experiments to analyze the effectiveness of each module.

Our main contributions are summarized as follows:
\begin{itemize}
  \item We propose $FG$ score, a new fine-grained evaluate metric for better distinguish the predicted keyphrases.
  \item Base on our evaluation metric, we propose a two-stage reinforcement learning method to optimize the model throughout a better direction.
  \item We train a BERT-based scorer whose corpus come from previous training. The scorer can effectively perceive the similarity of two keyphrases on semantic level.
  \item We conduct exhaustive experiments and analysis to show the effectively of our proposed $FG$ metric.
\end{itemize}

\section{Related Work}

In this section, we briefly introduce keyphrase generation models and evaluation metrics.

\subsection{Keyphrase Generation Models}

In KG task, keyphrases can be categorized into two types: {\it present} and {\it absent}, depending on whether it can be found in the source document or not. In recent years, end to end neural model has been widely-used in generating both present and absent keyphrases. ~\citet{DBLP:conf/acl/MengZHHBC17} introduced CopyRNN, which consists of an attentional encoder-decoder model~\cite{DBLP:conf/emnlp/LuongPM15} and a copy mechanism~\cite{DBLP:conf/acl/GuLLL16}. After that, relevant works are mainly based on the sequence-to-sequence framework. More recently, ~\citet{DBLP:conf/emnlp/ChenZ0YL18} leverages the coverage~\cite{DBLP:conf/acl/TuLLLL16} mechanism to incorporate the correlation among keyphrases, ~\citet{DBLP:conf/aaai/ChenGZKL19} enrich the generating stage by utilizing title information, and ~\citet{DBLP:conf/acl/ChenCLK20} proposed hierarchical decoding for better generating keyphrases. In addition, there are some works focus on keyphrase diversity~\cite{DBLP:conf/acl/YeGL0Z20}, selections~\cite{DBLP:conf/naacl/ZhaoBWWHZ21}, different module structure~\cite{xu2021nlpcc-searching-keyphrase}, or linguistic constraints~\cite{DBLP:conf/acl/ZhaoZ19}.

\subsection{Keyphrase Generation Metrics}

 Different to other generation tasks that need to generate long sequences, KG task only need to generate some short keyphrases, which means n-gram-based metrics (e.g., ROUGE~\cite{lin2004rouge}, BLEU~\cite{papineni2002bleu}) may not suitable for evaluations. Therefore, $F_1@5$~\cite{DBLP:conf/acl/MengZHHBC17} and $F_1@M$~\cite{DBLP:conf/acl/YuanWMTBHT20} are used to evaluate the keyphrases which is predicted by models. This evaluation score is also used as an adaptive reward to improve the performance through reinforcement learning approach~\cite{DBLP:conf/acl/ChanCWK19}.

\section{Methodology}
\subsection{Problem Definition}
In this section, we will briefly define the keyphrase generation problem. Given a source document $\mathbf{x}$, the objective is to predict a set of keyphrases $\cP=\{p_1,p_2,\dots,p_{|\cP|}\}$ to maximum match the ground-truth keyphrases $\mathcal{Y}=\{y_1,y_2,\dots,y_{|\mathcal{Y}|}\}$, where $|\cP|$ and $|\cY|$ are the number of the predicted keyphrases and the number of ground truth keyphrases respectively. Both source document $\mathbf{x} = [x_{1},...,x_{|\mathbf{x}|}]$ and a keyphrase in the set of target keyphrases $y_{i} = [y_{i, 1},...,y_{i, |y_{i}|}]$ are words sequences, where $|\mathbf{x}|$ and $|y_{i}|$ represent the length of source sequence $\mathbf{x}$ and the $i$-th keyphrase sequence $\mathbf{y}_{i}$, respectively.

\subsection{Seq2Seq Model with Minimizing Negative Log Likelihood Training}
In this section, we descibe the Seq2Seq model with attention~\cite{DBLP:conf/emnlp/LuongPM15} and copy mechanism~\cite{DBLP:conf/acl/GuLLL16}, which is our backbone model.

\paragraph{Encoder-Decoder Model with Attention}
We first convert the source document $\mathbf{x} = [x_1,x_2,...,x_{|\bx|}]$ to continuous embedding vectors $\mathbf{e} = [e_1,e_2,...,e_{|{\bx}|}]$. Then we adopt a bi-directional Gated-Recurrent Unit (GRU)~\cite{DBLP:conf/emnlp/ChoMGBBSB14} as the encoder to obtain the hidden state $\mathbf{H} = \mathrm{Encoder}(\mathbf{e})$.

Then another GRU is adopted as the decoder. At the step $t$, we compute the decoding hidden state $\bS_t$ as folow:
\begin{align}
\mathbf{S}_t = \mathrm{Decoder}(\mathbf{e}_{t-1}, \mathbf{s}_{t-1})
\end{align}

In addition, we incorporate the attention mechanism \cite{DBLP:conf/emnlp/LuongPM15}  to compute the contextual vector $\mathbf{u}$ which represents the whole source document at step $t$:
\begin{align}
    \mathbf{u}_t = \sum_{j=1}^{T}{\alpha_{tj}\mathbf{H}_j}
 \end{align}
 where $\alpha_{tj}$ represents the correlation between the source document at position $j$ and the output of the decoder at step $t$.

\paragraph{Copy Mechanism}
Because there are a certain number rare words in the document, traditional Seq2Seq models perform pooly when predicting these rare words. Thus, we introduce the copy mechanism \cite{DBLP:conf/acl/GuLLL16} to alleviate the out-of-vocabulary (OOV) problem. The probability of producing a token contains two parts: the probability for generation $p_{g}$ and probability for copy mechanism $p_{c}$.  $p_{g}$ is estimated by a standard language model based on the global vocabulary, and $p_{c}$ is estimated by the copy distribution based on local vocabulary which only contain one case. The definition of $p_{c}$ is:
\begin{align}
p_{c}\left(y_{i,t}|y_{i,<t}, \mathbf{H}\right) &= \frac{1}{Z} \sum_{j:x_{j}=y_{i,t}}{e^{\omega(x_j)}}, y_{i,t} \in \chi
\end{align}
where $\chi$ represents the set of all rare words in the source document and $Z$ is used for normalization.

\paragraph{Minimizing Negative Log Likelihood Training}
Finally, we train all parameters in the model $\theta$ by minimizing the negative log likelihood loss:
\begin{align}
\mathcal{L}(\theta) = -\sum{\mathrm{log}P\left(y_{i,t}|y_{i,<t}, \mathbf{H}\right)}
 \end{align}

\subsection{Fine-Grained Score}
\label{sec-fg-score}

Because traditional KG methods only care predictions on phrase-level in evaluate stage, they ignore both information on token-level and instance-level. Not only in the traditional Seq2Seq model, there are also in RL training. The environment also calculates the reward (recall or $F_1$-score) only in phrase-level, which ignores the overall performance of prediction. Due to this problem, the training process may go in the wrong direction (e.g. Model will give a zero score for a phrase that has more than half right). Thus, we propose a new metric: \textbf{F}ine \textbf{G}rained Score ($FG$), which considers both token-level and instance-level information. It is divided into the following four parts.

\subsubsection{Phrase Similarity on Token Level}
\label{sec-phrase-similarity-token-level}

For the comprehensive calculation later, we first compute the similarity score between a  predicted phrase and a ground-truth phrase on token-level. We use edit distance and token-level $F_1$ score as our metric.

Obviously, in order to get token-level similarity, Guaranteed the correctness of phrase on token-level is important. So token-level $F_1$ score is necessary. Given a predicted phrase $p_i\in \cP$, we use $\textbf{F}_1(p_i, y_j)$ represent the score for $i$-th predicted phrase and $j$-th ground-truth phrase.

Because the order of the words in a phrase is also important, we introduce the edit distance to measure the sequential difference between the two phrases. Particularly, the edit distance $\mathbf{ED}(p_i,y_j)$ denotes how many times should modify $p_i$ to $y_j$ at least, where one time only can modify one word and modify operation only contains three operations: {\textbf{add}, \textbf{delete}, \textbf{change}}. We use dynamic programming to calculate the edit distance as follow:

\begin{align}
  D_{k}^{m}=\left\{
             \begin{array}{l}
             \mathrm{min}(D_{k-1}^{m-1},D_{k}^{m-1}+1,D_{k-1}^{m}+1)\\
             \text{if}\; p_{i,k} = y_{j,m}  \\\\
             \mathrm{min}(D_{k-1}^{m-1}+1,D_{k}^{m-1}+1,D_{k-1}^{m}+1)\\
             \text{if}\; p_{i,k} \ne y_{j,m}  \\
             \end{array}
             \right.
\end{align}
where $D_{k}^{m}$ denotes minimum number of modifications for transforming first $k$ token in $p_i$ to first $m$ token in $y_j$. $k \in [1, |p_i|]$ and $m \in [1, |y_j|]$.
Because the more modifications there are, the less similar the two sequences are, the $\mathbf{ED}(p_i,y_j)$ score can be formulated as follow:
\begin{align}
    \mathbf{ED}(p_i,y_j) = 1 - \frac{D_{|p_i|}^{|y_j|}}{\mathrm{max}\{|p_i|, |y_j|\}}
\end{align}

And for a instance ($\mathbf{x}$, $\mathcal{Y}$, $\mathcal{P}$), we compute score list $\mathbf{scoreL}$ as follow:
\begin{align} \label{scoreL}
  \mathbf{score}\bL_i=\mathop{\max}_{y_j}\{\frac{\mathbf{ED}(p_i,y_j) + \textbf{F}_1(p_i, y_j))}{2}\},
\end{align}
where $\mathbf{F}_1$ is token-level $F_1$ score. $i \in [1, |\mathcal{P}|]$ and $j \in [1, |\mathcal{Y}|]$. And we use a maximum-match score to a particularly predicted phrase.

\subsubsection{Global Generation Quality on Instance Level}
\label{sec-global-generation-quality-instance-level}

In Section~\ref{sec-phrase-similarity-token-level}, we proposed a method to compute the phrase similarity on token level. In this section, we will further consider the generation quality on instance level.

There are many factors can influent the global generation quality, but we select the most representative factors: diversity and the prediction quantities. Therefore, we use a {\bf Repetition Rate Penalty} and {\bf Generation Quantity Penalty} for the $FG$ score, which is shown in Algorithm~\ref{algorithm}.

\begin{algorithm}[ht]
\begin{algorithmic}[1]
\REQUIRE ~~\\
The set of ground-truth keyphrases, $\mathcal{Y}$;\\
The set of predicted keyphrases $\mathcal{P}$;\\
The score list of prediction, $\textbf{score}\bL$
\ENSURE ~~\\
reward for an instance;
\STATE $//$ Repetition rate penalty
\STATE initial two dicts $dictY$ and $dictP$
\FORALL{keyphrase $\mathbf{y}_i$ in $\mathcal{Y}$}
\FORALL{word $y_{ij}$ in $\mathbf{y}_i$}
\STATE $dictY[y_{ij}] = dictY[y_{ij}] + 1$
\ENDFOR
\ENDFOR
\STATE reverse sort $\mathcal{P}$ and $\mathbf{score}\bL$ by key $\mathbf{score}\bL$
\FOR{$i=0$; $i<|\mathcal{P}|$; $i++$}
\FORALL{word $p_{ij}$ in $p_i$}
\IF{$p_{ij}$ in $dictY$}
\STATE $dictP[p_{ij}]++$
\IF{$dictP[p_{ij}] > dictY[p_{ij}]$}
\STATE $\mathbf{score}\bL[i] = 0$
\ENDIF
\ENDIF
\ENDFOR
\ENDFOR
\STATE $finalscore = \frac{\mathrm{sum}(\mathbf{score}\bL)}{|\mathcal{P}|}$
\STATE $//$ Generatation quantity penalty
\STATE $corr = 1.0 - \frac{(|\mathcal{Y}| - |\mathcal{P}|)^2}{\mathrm{max}(|\mathcal{Y}|, |\mathcal{P}|)^2}$
\STATE $finalscore = finalscore * corr$
\RETURN $finalscore$
\end{algorithmic}
\caption{Global Generation Quality Penalty}
\label{algorithm}
\end{algorithm}

The first factor is the repetition rate penalty. This operation means that there is a punishment if the model predicts similar keyphrases greater equal than twice, which can also reduce the duplication. We first count the words that appear in the ground truth. Then we sort the prediction and the score list in reverse according to the score. After that we iterate the prediction list and count the words that appear in the predictions. Once a word appears in the predictions twice more than that in the ground truth, we claim this token ``repetitive''. Based on this, the corresponding phrase is labelled as invaild. Lastly we will compute an average score of all phrases as a normalization, which can be used to represent the score of the corresponding instance.

The second factor is the generation quantities. The model will obtain the highest score if it predicts only one most simple phrase because it is an exact match result in most cases. Therefore, we add a generation coefficient to solve this problem.

\subsection{Continuous Scorer}
\label{sec-bert-scorer}

In fact, although the $FG$ metric includes the token-level and instance-level information for keyphrase, deeper semantic information is not considered. As the introduction says, when ``{\it natural language processing}'' is ground truth,  our $FG$ metric will give ``{\it natural language understanding}'' and ``{\it natural language generation}'' a same score. But ``{\it natural language understanding}'' and ``{\it natural language generation}'' have different semantics. In order to solve this problem, we incorporate pre-train model (e.g., BERT) to train a continuous scorer which denotes the similarity of two keyphrases.

Because many tuples ($p_i$, $y_j$, $\mathbf{scoreL}_i$)  are generated when we compute the $FG$ score, we screen portions as training corpus for BERT.
We concatenate the $p_i$ and $y_j$ as (\texttt{[CLS]} $p_i$ \texttt{[SEP]} $y_j$ \texttt{[SEP]}) to a sequence as input for BERT scorer, where \texttt{[CLS]} and \texttt{[SEP]} is the same as the vanilla BERT~\cite{devlin-etal-2019-bert}. In the training stage, $\mathbf{scoreL}_i$ score is used as the supervised target.

After get the BERT scorer, we can easily evaluate the similarity of two keyphrase. Similar to the Eq ~\eqref{scoreL}, for a instance ($\mathbf{x}$, $\mathcal{Y}$, $\mathcal{P}$), we compute BERT-based score list $\mathbf{scoreLB}$ as follow:

\begin{align} \label{scoreLB}
    \mathbf{score}\bL\bB_i = \mathop{\max}_{y_j}\{\mathrm{BERT}(p_i, y_j)\}.
\end{align}
where $i \in [1, \mathcal{P}]$ and $j \in [1, \mathcal{Y}]$.
Finally, we also put $\mathbf{scoreLB}_i$ into Algorithm~\ref{algorithm}
to compute finally BERT-based score (also called $FB$ score).

\subsection{Reinforcement Learning}

In this section, we will briefly describe our proposed two-stage reinforcement learning method.

\subsubsection{Vanilla RL Training}
Reinforcement learning has been widely applied to text generation tasks,  such as machine translation \cite{DBLP:conf/emnlp/WuTQLL18}, summarization \cite{DBLP:conf/naacl/NarayanCL18}, because it can train the model towards a non-differentiable reward. \citet{DBLP:conf/acl/ChanCWK19} incorporate reinforce algorithm to  optimize the Seq2Seq model with an adaptive reward function.
They formulate keyphrase generation as follow. At the time step $t = 1, \dots, T$, the agent produces an action (token) $\widehat{y}_t$ sampled from the policy (language model) $P(\widehat{y}_t|\widehat{y}_{<t})$, where $\widehat{y}_{<t}$ represent the sequence generated before step $t$. After generated $t$-th tokens, the environment $\widehat{s}_{t}$ will gives a reward $r_t(\widehat{y}_{<=t}, \mathcal{Y})$ to the agent and updates the next step with a new state $\widehat{s}_{t+1} = (\widehat{y}_{<=t}, \mathbf{x}, \mathcal{Y})$. We repeat the above operations until generated all token.
Typically, the recall score or the $F_1$ score are used as the reward function.

\subsubsection{Two-Stage RL Training}
\label{two-stage}

In the vanilla RL training, the reward is polarized in the phrase level: one for an exact match prediction and zero for other situations, which means a partial match phrase receives the same reward as an exact mismatch phrase. In order to help to recognize these partial match phrases during the training stage, we propose a two-stage RL training method. In the first stage, we use our new metric ($FG$ score or $FB$ score) as a reward to train the model. Then we apply the vanilla RL (using $F_1$ score) training as the second training stage. The whole RL training technique is similar to \citet{DBLP:conf/acl/ChanCWK19}, while we re-write the reward function.

\begin{table*}[tp]
	\centering\small
	\scalebox{0.9}{
	\begin{tabular}{l|c|c|c|c|c|c|c|c|c}
        \toprule
		\multirow{2}{*}{\textbf{Model}} & \multicolumn{3}{c|}{\textbf{Inspec}} &   \multicolumn{3}{c|}{\textbf{Krapivin}}  & \multicolumn{3}{c}{\textbf{KP20k}} \\
		& \multicolumn{1}{c}{$F_{1}@M$} &  \multicolumn{1}{c}{$F_{1}@5$} &  \multicolumn{1}{c|}{$\color{red}FG$}
		&\multicolumn{1}{c}{$F_{1}@M$} &  \multicolumn{1}{c}{$F_{1}@5$} &
		\multicolumn{1}{c|}{$\color{red}FG$}
		&\multicolumn{1}{c}{$F_{1}@M$} &  \multicolumn{1}{c}{$F_{1}@5$} &
		\multicolumn{1}{c}{$\color{red}FG$} \\
        \midrule
		catSeq\cite{DBLP:conf/acl/YuanWMTBHT20} & 0.262 & 0.225  & 0.381 & 0.354 & 0.269 & 0.352 &0.367 & 0.291 & 0.371 \\
		catSeqD\cite{DBLP:conf/acl/YuanWMTBHT20}  & 0.263 & 0.219  & 0.385 & 0.349 & 0.264 & 0.350 &0.363 & 0.285 & 0.369 \\
		catSeqCorr\cite{DBLP:conf/emnlp/ChenZ0YL18}  & 0.269 & 0.227 & 0.391 &0.349 & 0.265 & 0.360 &0.365 & 0.289 & 0.374\\
		catSeqTG\cite{DBLP:conf/aaai/ChenGZKL19}  & 0.270 & 0.229 & 0.391 &0.366 & 0.282 & 0.344 &0.366 & 0.292 & 0.369 \\
    SenSeNet\cite{luo2020sensenet} & 0.284 & 0.242 & 0.393 & 0.354 & 0.279 & 0.355 & 0.370 & 0.296 & 0.373 \\
		ExHiRD-h\cite{DBLP:conf/acl/ChenCLK20} & 0.291 & \underline{0.253} & \underline{0.395} & 0.347 & 0.286 & 0.354 & 0.374 & 0.311 & 0.375 \\
    \midrule
    \multicolumn{10}{l}{Utilizing RL~\cite{DBLP:conf/acl/ChanCWK19}}\\
    \midrule
    catSeq+RL($F_1$)     & 0.300 & 0.250 & 0.382 & 0.362 & 0.287 & 0.360 & 0.383 & 0.310 & 0.369 \\
    catSeqD+RL($F_1$)    & 0.292 & 0.242 & 0.380 & 0.360 & 0.282 & 0.357 & 0.379 & 0.305 & \underline{0.377} \\
      catSeqCorr+RL($F_1$) & 0.291 & 0.240 & 0.392 & \underline{0.369} & 0.286 & \underline{0.376} & 0.382 & 0.308 & \underline{0.377} \\
    catSeqTG+RL($F_1$)   & \underline{0.301} & \underline{0.253} & 0.389 & \underline{0.369} & \underline{0.300} & 0.344 & \underline{0.386} & \underline{0.321} & 0.370 \\
    \midrule
    \midrule
    \multicolumn{10}{l}{Ours}\\
		\midrule
		catSeq*+RL($FG$)            & 0.252 & 0.201 & 0.460 & 0.359 & 0.228 & 0.413 & 0.365 & 0.290 & 0.440 \\
		catSeq*+RL($FB$)          & 0.254 & 0.200 & \textbf{0.463} & 0.354 & 0.230 & \textbf{0.416} & 0.366 & 0.291 & \textbf{0.444} \\
		catSeq*+2RL($FG$)           & 0.308 & 0.266 & 0.425 & \textbf{0.375} & 0.304 & 0.389 & 0.391 & 0.327 & 0.381 \\
	    catSeq*+2RL($FB$)         & \textbf{0.310} & \textbf{0.267} & 0.430 & 0.374 & \textbf{0.305} & 0.390 & \textbf{0.392} & \textbf{0.330} & 0.383 \\
		\bottomrule
	\end{tabular}
	}
	\caption{Result of present keyphrase prediction on three datasets. ``RL'' denotes that a model is trained by one-stage reinforcement training. ``2RL'' denotes that a model is trained by two-stage RL training. The notation in parentheses denotes the reward function in first RL training stage. All second reward function in two-stage RL training is $F_1$ score. ``catSeq*'' represents that we select the best model of four different catSeq-based baseline models. $FB$ indicates that the reward is computed by the continuous BERT scorer. The underline numbers represent the best result in previous work. $\color{red}FG$ is the metric we propose.}
	\label{tab:present}
\end{table*}

\section{Experiment}
\subsection{Dataset}
We evaluate our model on three public scientific KG dataset, including \textbf{Inspec} \cite{DBLP:conf/acl/HulthM06}, \textbf{Krapivin} \cite{krapivin2009large}, \textbf{KP20k} \cite{DBLP:conf/acl/MengZHHBC17}. Each case from these datasets
consists of the title, abstract, and a set of keyphrases. Following the previous work \cite{DBLP:conf/acl/ChenCLK20}, we concatenate the title and abstract as input document, and use the set of keyphrases as labels. The same as the previous works above, we use the largest dataset, \textbf{KP20k}, to train the model, and use all datasets to evaluate the performance of our model. After same data pre-processing as \citet{DBLP:conf/acl/ChanCWK19}, \textbf{KP20k} dataset contains 509,818 training samples, 20,000 validation samples, and 20,000 testing samples.

\subsection{Evaluation Metrics}
Most previous work \cite{DBLP:conf/acl/MengZHHBC17,DBLP:conf/emnlp/ChenZ0YL18, DBLP:conf/aaai/ChenGZKL19} cutoff top $k$ (which $k$ is a fixed number) predicted keyphrases to calculate metrics such as $F_1@5$ and $F_1@10$.  Due to the different number of keyphrases in different samples, \citet{DBLP:conf/acl/YuanWMTBHT20} propose a new evaluation metric, $F_1@M$, which compares all keyphrases predicted with the ground-truth and compute the $F_1$ score. We evaluate the performance of our model using three different metrics, $F_1@5$, $F_1@M$, and {\color{red}$FG$} (ours). After computing every samples' scores, we apply marco average to aggregate the evaluation scores. The same as \citet{DBLP:conf/acl/ChanCWK19}, we append random wrong keyphrases to prediction until it reaches five or more, because our method generates diverse keyphrases that usually less than five predictions.

\subsection{Baseline Model}
Following the name set of the previous works\cite{DBLP:conf/acl/ChanCWK19, DBLP:conf/acl/ChenCLK20}, we use four generative model trained under minimize the negative log likelihood loss, include \textbf{catSeq}\cite{DBLP:conf/acl/YuanWMTBHT20}, \textbf{catSeqD}\cite{DBLP:conf/acl/YuanWMTBHT20}, \textbf{catSeqCorr}\cite{DBLP:conf/emnlp/ChenZ0YL18}, \textbf{catSeqTG}\cite{DBLP:conf/aaai/ChenGZKL19}, \textbf{ExHiRD-h}\cite{DBLP:conf/acl/ChenCLK20}. Because reinforcement learning is applied to our method, we also compare four reinforced model \cite{DBLP:conf/acl/ChanCWK19} include \textbf{catSeq+RL}, \textbf{catSeqD+RL}, \textbf{catSeqCorr+RL}, \textbf{catSeqTG+RL}. Each reinforced model is correspond to previous model applied RL approach. In this paper, our RL framework trains four models for comparison:
\begin{itemize}
    \item \textbf{catSeq*+RL($FG$)} and \textbf{catSeq*+RL($FB$)}  denotes that one-stage reinforcement learning training with $FG$-score reward or BERT-based reward.

    \item \textbf{catSeq*+2RL($FG$)} and \textbf{catSeq*+2RL($FB$)} denotes that two-stage RL training. Two methods use $FG$-score and BERT-based reward in first stage respectively, and both use $F_1$ reward in second score which is same as \citet{DBLP:conf/acl/ChanCWK19}.
\end{itemize}

\begin{table*}[h]
	\centering
  \small
  \tabcolsep 8.0pt
	\scalebox{0.9}{
	\begin{tabular}{l|c|c|c|c|c|c}
        \toprule
		\multirow{2}{*}{\textbf{Model}} &
		\multicolumn{3}{c|}{\textbf{Phrase-level Result}} &
		\multicolumn{3}{c}{\textbf{Token-level Result}}
		\\
		& \multicolumn{1}{c}{$F_1@M$} &  \multicolumn{1}{c}{$F_1@5$} & \multicolumn{1}{c|}{\color{red} $FG$} &
		\multicolumn{1}{c}{$tF$} & \multicolumn{1}{c}{$tP$} &
		\multicolumn{1}{c}{$tR$}
		\\
        \midrule
		catSeq*+2RL($FG$) & \textbf{0.391} & \textbf{0.330} & \textbf{0.383} & 0.494 & 0.493 & 0.495 \\
		\multicolumn{1}{r|}{w/o ED} & 0.387 & 0.325 & 0.370 & 0.494 & 0.491 & \textbf{0.498} \\
		\multicolumn{1}{r|}{w/o TF} & 0.389 & 0.327 & 0.372 & 0.485 & 0.483 & 0.487 \\
		\multicolumn{1}{r|}{w/o RRP}  & 0.390 & 0.328 & 0.375 & \textbf{0.497} & \textbf{0.494} & 0.500 \\
		\multicolumn{1}{r|}{w/o GNP}  & 0.388 & 0.320 & 0.372 & 0.489 & 0.493 & 0.486 \\
        \bottomrule
	\end{tabular}
	}
	\caption{Ablation study of catSeq*+2RL($FG$) on \textbf{KP20k} dataset. ED means Edit Distance, TF means Token-level $F_1$ score (see Section~\ref{sec-phrase-similarity-token-level}), RRP means Repetition Rate Penalty, GNP means Generated Number Penalty (see Section~\ref{sec-global-generation-quality-instance-level}). ``w/o'' means ``without''. $tF$, $tP$, $tR$ means token-level metric. }
	\label{tab:ablation study}
\end{table*}

\section{Result and Analysis}
\subsection{Present Keyphrase Prediction}
In this section, we evaluate the performance of our models on present keyphrase predictions using three different metrics, $F_1@M$, $F_1@5$, and {\color{red}$FG$}, respectively. Table \ref{tab:present}  shows the result of all baseline models and our proposed four models.
From the result, we summarized our observations as follow:

(1) Our proposed methods achieve the state-of-the-art result on KG generation, which proves that it is necessary to deal with the semantic similarities between predictions and targets.

(2) In the phrase level, the reward returned by the vanilla RL method (with $F_1$ score) is polarized. Assuming that there are two partial match predictions in the baseline model (catSeq*), one of them may change into an exact match keyphrase while another may change into an exact mismatch keyphrase after the vanilla RL method. This phenomenon will increase the $F_1$ score, but only a similar $FG$ score can be obtained. Therefore, the vanilla RL method hardly improves the quality of generation, although it improves the $F_1@5$ and $F_1@M$ score.

(3) We observe that the one-stage RL training (catSeq*+RL($FG$)) induces the performance drop on both $F_1@M$ and $F_1@5$, especially on $F_1@5$, but it improves the performance on $FG$. The reason is that the number of predicted keyphrases is less than vanilla RL training. We predict 3.2 present keyphrases on average, and the vanilla RL training predicts 3.8 when ground truth is 3.3. We conclude that the number of our predictions is more reasonable comparing with the vanilla RL methods.

(4) Models with two-stage RL training far outperform those with only one-stage RL training on $F_1@M$ and $F_1@5$ metrics. Moreover, it shows that the vanilla RL training with $F_1$ score can effectively improve $F_1@5$ and $F_1@M$ after first stage training because first stage training improves the token-level quality of prediction.

(5) We observe that using BERT as a reward scorer makes the models perform better than using $FG$, indicating that the reward score produced by BERT is usually more accurate.

\begin{figure*}[!ht]
\centering
\includegraphics[width=0.98\linewidth]{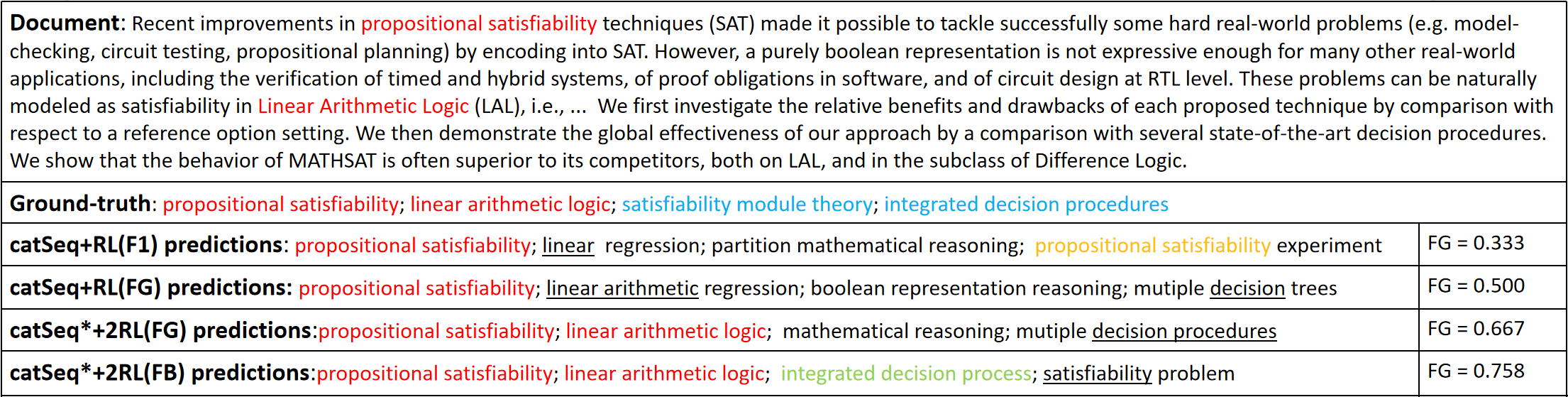}
\caption{Case study for catSeq+RL($F_1$), catSeq+RL($FG$), catSeq*+2RL($FG$) and catSeq*+2RL($FB$). The red words represent the present keyphrases, the blue words represent the absent keyphrase. The green words represent the synonym with ground truth. The yellow words represent the duplicate part of a keyphrase. The underlined words represent correctly words on token-level. }
\label{fig:case}
\end{figure*}

\begin{figure*}[t]
\centering
  \begin{subfigure}{0.42\linewidth}
 \resizebox{0.75\columnwidth}{!}{%
    \begin{tikzpicture}
\begin{axis}[
  ybar,
  grid=both,
  every tick label/.append style={scale=0.85},
  ymax=14200,
  ymin=0.0,
  xmin=0.3,
  xmax=4.5,
  axis y line*=left,
  axis x line*=bottom,
  xticklabels={[0.0\,0.4),[0.4\,0.7),[0.7\,1.0),[1.0\,)},xtick={1,2,3,4},
  ytick={0,2000,4000,6000,8000,10000,12000,14000},
  ylabel= \#instances,
  xlabel=$FG$ score,
  legend style={nodes={scale=0.8, transform shape},at={(0.5,1)},anchor=north}]
\addplot
	coordinates {(1,11917)(2,6018)(3,1786)(4,279)};
\addplot
	coordinates {(1,9176)(2,6630)(3,4203)(4,796)};
\addplot
	coordinates {(1,11363)(2,6291)(3,1990)(4,356)};
\legend{catSeq+RL($F1$), catSeq*+RL($FG$), catSeq*+2RL($FG$)}
\end{axis}
\end{tikzpicture}
}
\caption{Number distribution histogram of $FG$ score}\label{fig:d1}
  \end{subfigure}
\begin{subfigure}{0.42\linewidth}
\resizebox{0.85\columnwidth}{!}{%
  \begin{tikzpicture}
\begin{axis}[
  ybar,
  grid=both,
  every tick label/.append style={scale=0.85},
  ymax=0.46,
  ymin=0.0,
  xmin=-0.5,
  xmax=10.5,
  axis y line*=left,
  axis x line*=bottom,
  xticklabels={[\,0.0),[0.0\,0.4),[0.4\,0.7),[0.7\,1.0),[1.0\,)},xtick={1,3,5,7,9},
  ytick={0,0.05,0.1,0.15,0.2,0.25,0.3,0.35,0.4,0.45},
  ylabel=\#keyphrases / \#total keyphrases,
  xlabel=token-level $F_1$ score,
  legend style={nodes={scale=0.8, transform shape},at={(0.5,1)},anchor=north}]
\addplot
	coordinates {(1,0.414)(3,0.041)(5,0.141)(7,0.062)(9,0.341)};
\addplot
	coordinates {(1,0.314)(3,0.036)(5,0.147)(7,0.083)(9,0.419)};
\addplot
	coordinates {(1,0.400)(3,0.04)(5,0.143)(7,0.061)(9,0.352)};
\legend{catSeq+RL($F1$), catSeq*+RL($FG$), catSeq*+2RL($FG$)}
\end{axis}
\end{tikzpicture}
}
\caption{Proportion distribution histogram of token-level $F_1$ score }\label{fig:d2}
\end{subfigure}
\caption{The number distribution of the $FG$ and the proportion distribution of token-level $F_1$ score on test dataset by three different training process. catSeq+RL($F_1$) denotes that one-stage RL training with $F_1$-score reward. catSeq*+RL($FG$) denotes that one-stage RL training with $FG$-score reward. catSeq*+2RL($FG$) denotes that two-stage RL training with $FG$ and $F_1$-score reward.}\label{fig:total_d}
\end{figure*}
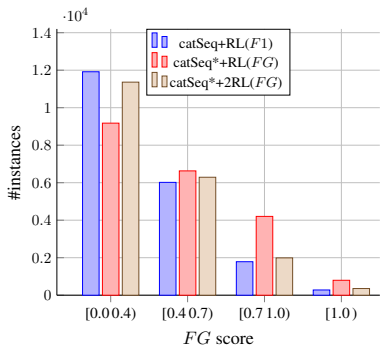
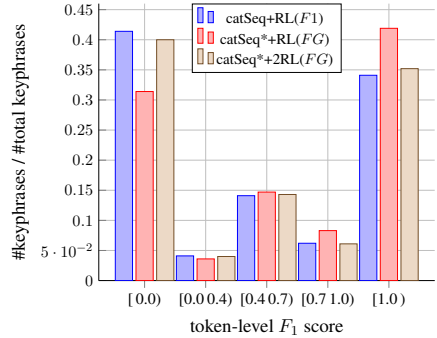

\subsection{Ablation Study}
To further examine the benefits that each component of the $FG$ score brings to the performance, we conduct an ablation study on the \textbf{catSeq*+2RL($FG$)} model. Our proposed methods are evaluated on the largest dataset \textbf{KP20k}. The results are shown in Table \ref{tab:ablation study}.

First, removal of edit distance score (w/o ED) does not affect model's performance on token-level but leads to performance drop most on phrase-level. Thus, it proves that edit distance is the most crucial in $FG$ scores. Moreover, after we get rid of token-level $F_1$ (w/o TF), we observe that the phrase-level performance does not decrease much, but token-level performance decrease much. Therefore, we prove the effectiveness of token-level $F_1$ for token-level quality.

Compared with \textbf{catSeq*+2RL($FG$)}, removal of the repetition rate penalty (w/o RRP) will cause the performance drop consistently on phrase-level, which indicates that RRP has a great effect on phrase-level $F_1@5$. Furthermore, for token-level results, we observe that the token-level recall and token-level $F_1$ score decreases somewhat, but token-level precision gets a promotion.
From predicted results, we also obtain some observations when the lack of repetition rate penalty. There are a large number of keyphrase such as ``{\it natural processing}'', ``{\it natural language}'', ``{\it natural natural natural}'', when the ground-truth keyphrase is ``{\it natural language processing}''.  The situation leads to high token-level accuracy but low overall performance.

Finally, removal of generated number penalty (w/o GNP) will mainly cause the phrase-level $F_1@5$ to go down. We find that model tends to generate a small number of keyphrases as the predicted results because generating multiple keyphrases will reduce the reward.
According to the definition of $F_1@5$, if the model can not generate enough five keyphrases, we should randomly add a mistake keyphrase to five. Thus, if we generate more keyphrases appropriately, $F_1@5$ will definitely get a boost.
So in this situation, $F_1@5$ will decrease a lot.
From what has been discussed above, all the modules in the $FG$ score have their contribution.

\subsection{Case Study}
To better understand what benefits our proposed method brings, we present a case study with a document sample.
As shown in the Figure \ref{fig:case}, we compare the predictions  generated by vanilla RL model (catSeq+RL($F_1$)), two-stage RL training with $FG$ score (catSeq*+2RL($FG$)) and two-stage RL training with BERT scorer (catSeq*+2RL($FB$)) on a same document sample. Overall, our two approaches have improved relative to the baseline model on $FG$ scorer, and it shows that our overall generation quality has been improved.

From the case, we have three observations: \textbf{First},  catSeq*+2RL($FG$) and catSeq*+2RL($FB$) correctly predict the keyphrase ``{\it linear arithmetic logic}'' while catSeq+RL($F_1$) predicts a ``{\it linear regression}'' which gets only one word right. It indicates that our two methods can improve the prediction quality on token-level and then finally improve the performance on phrase-level. \textbf{Second}, catSeq+RL($F_1$) predicts two similarly keyphrases ``{\it proposition satisfiability}'' and ``{\it proposition satisfiability experiment}'', which our two methods do not. It fully demonstrates that our repetitive punishment plays an important role, which makes the predictions become diverse. \textbf{Third}, catSeq*+2RL($FB$) generates a keyphrase ``{\it integrated decision process}'', which is synonym for ground truth ``{\it integrated decision procedures}''. It indicates that the BERT scorer can effectively perceive the semantics of keyphrases, which guides the training process of reinforcement learning.

\subsection{Generative Quality Analysis}

In this section, we analyze the prediction quality generated by vanilla RL model (catSeq+RL($F_1$)),  one-stage RL training with $FG$ score (catSeq*+RL($FG$)) and two-stage RL training with $FG$ score (catSeq*+2RL(FG)) on instance-level and token-level respectively.  We both divided the $FG$ score and token-level $F_1$ score into five parts. We conduct detailed analysis in the following.

In Figure~\ref{fig:d1},  we use the distribution of $FG$ scores to analyze the generation quality on instance-level. By comparing the distribution of catSeq+RL($F_1$) and catSeq*+RL($FG$),  we find that catSeq*+RL($F_1$) has a larger proportion when $FG$ score is low and catSeq*+RL($FG$) has a larger proportion when $FG$ score is high. It shows that using the $FG$ score as a reward can improve the overall quality of predictions. Especially when score = 1.0 (which means all keyphrase is correctly in this instance), the number of catSeq*+RL($FG$) is nearly three times as large as catSeq+RL($F_1$). When comparing catSeq*+2RL($FG$) with catSeq+RL($F_1$), we obtain the similar conclusion as before. It is proved that the overall quality of the generated keyphrases can be improved after the first stage of reinforcement learning training.

In Figure~\ref{fig:d2}, due to different number keyphrases predicted by the model, we use the distribution of the proportion of the token-level $F_1$ score to analyze the generation quality on token-level. By comparing the distribution of catSeq+RL($F_1$) and catSeq*+RL($FG$),  we find that catSeq*+RL($F_1$) has a larger proportion when token-level $F_1$ is low and catSeq*+RL($FG$) has a larger proportion when token-level $F_1$ is high. It indicates that the model can generate more keyphrases with more correct words throughout the reinforcement learning training with the $FG$ score. (e.g. When groud truth is ``{\it natural language processing}'', catSeq+RL($F_1$) generates ``{\it natural X X}'' and catSeq*+RL($FG$) generates ``{\it natural language X}''. ``{\it X}'' means the inaccuracy word).  This improvement also benefits to catSeq*+2RL($FG$).

\subsection{Human Evaluation for Continuous Scorer}

As shown in Section~\ref{sec-bert-scorer}, our continuous BERT-based scorer is an implicit and automatic. In this section we manually evaluate it to verify its effectiveness. We randomly selected 1000 pairs of matching predicted and ground-truth keyphrases in the training of reinforcement learning with BERT-based rewards and save a BERT score at the same time. Especially, we do not select the keyphrase pairs whose score is below to 0.05 or above to 0.95, because these pairs are either completely unrelated or exactly the same.
We randomly divide the data into five samples and ask five different people to rate each pair of keyphrases (Scores range: 0.0, 0.1, ... , 0.9, 1.0). Both of the annotators have no less than a bachelor degree, which have the enough ability of evaluating the quality of model predictions. Then we used \textbf{Pearson} correlation coefficient and \textbf{Spearman} correlation coefficient to measure the effect of the BERT Scorer. The human evaluation results are shown in Table~\ref{human}. From the results, we can conclude that the scorer produced by BERT has high quality, and hence, it can act as a helpful signal during our training process.

\begin{table}
\centering
\small
    \scalebox{1.0}{
    \begin{tabular}{c| c c}
    \toprule
    Annotator & Pearson & Spearman\\
    \midrule
    People 1 & 0.894 & 0.884 \\
    People 2 & 0.881 & 0.867 \\
    People 3 & 0.874 & 0.856 \\
    People 4 & 0.889 & 0.873 \\
    People 5 & 0.883 & 0.875 \\
    \midrule
    Total & 0.884 & 0.870 \\
    \bottomrule
    \end{tabular}
    }
    \caption{The results of manually evaluation on Pearson and Spearson correlation coefficient.}
    \label{human}
\end{table}

\section{Conclusion}
In this paper, we utilize a two-stage reinforcement learning training framework with a fine-grained evaluation metric. We propose the $FG$-score or the continuous BERT-score as the reward in the first-stage training, which improves the generation quality on token-level and then beneficial to the second-stage training. Experiments on KG benchmarks show the effectiveness of our proposed method, and then we also demonstrated the contribution of each module in the $FG$ function. In addition, we evaluate the performance of BERT-based scorer manually. In future work, we will consider improving the training of BERT scorer's performance and making the two-stage RL training more effective.

\section*{Acknowledgement}

We thank the anonymous reviewers for their helpful comments. We also thank Xiaoyu Xing for her valuable feedback about the paper presentation. This work was partially funded by China National Key R\&D Program (No. 2017YFB1002104), National Natural Science Foundation of China (No. 61976056, 62076069), Shanghai Municipal Science and Technology Major Project (No.2021SHZDZX0103).

\bibliographystyle{acl_natbib}
\bibliography{acl2021}

\end{document}